\documentclass[11pt]{article}
\usepackage{acl}
\usepackage{times}
\usepackage{latexsym}
\usepackage{xspace}
\usepackage[T1]{fontenc}
\usepackage[utf8]{inputenc}
\usepackage{microtype}
\usepackage{graphicx}
\usepackage{url}
\usepackage{amsmath}
\usepackage{amssymb}
\usepackage{array}
\usepackage{booktabs}
\usepackage{multirow}
\usepackage{makecell}
\usepackage{tabularx}
\usepackage{subcaption}
\usepackage{xcolor}
\usepackage{enumitem}
\usepackage{amsmath}
\usepackage{csquotes}
\usepackage{tcolorbox}
\tcbuselibrary{breakable}
\tcbuselibrary{skins}
\newcommand{\system}{\textsc{Concord}\xspace}
\newcolumntype{C}{>{\centering\arraybackslash}X}
\ifx\undefined\natexlab\DeclareRobustCommand{\natexlab}[1]{#1}\fi
\ifx\undefined\url\DeclareRobustCommand{\url[1]{\texttt{#1}}\fi
\ifx\undefined\expandafter\selectfont\DeclareRobustCommand{\selectfont}{}\fi

\title{Listening Alone, Understanding Together: Collaborative Context Recovery for Privacy-Aware AI}

\author{
Tanmay Srivastava \\
Stony Brook University \\
\texttt{tsrivastava@cs.stonybrook.edu}
\And
Amartya Basu \\
Stony Brook University \\
\texttt{ambasu@cs.stonybrook.edu}
\AND
Shubham Jain \\
Stony Brook University \\
\texttt{jain@cs.stonybrook.edu}
\And
Vaishnavi Ranganathan \\
Microsoft Research \\
\texttt{vnattar@microsoft.com}
}

\begin{document}
\maketitle
\begin{abstract}
We introduce~\system{}, a privacy-aware asynchronous assistant-to-assistant (A2A) framework that leverages collaboration between proactive speech-based AI. 
As agents evolve from reactive to always-listening assistants, they face a core privacy risk (of capturing non-consenting speakers), which makes their social deployment a challenge. 
To overcome this, we implement ~\system{}, which enforces owner-only speech capture via real-time speaker verification, producing a one-sided transcript that incurs missing context but preserves privacy.
We demonstrate that ~\system{} can safely recover necessary context through (1) spatio-temporal context resolution, (2) information gap detection, and (3) minimal A2A queries governed by a relationship-aware disclosure. Instead of hallucination-prone inferring,~\system{} treats context recovery as a negotiated safe exchange between assistants.
Across a multi-domain dialogue dataset,~\system{} achieves 91.4\% recall in gap detection, 96\% relationship classification accuracy, and 97\% true negative rate in privacy-sensitive disclosure decisions. By reframing always-listening AI as a coordination problem between privacy-preserving agents,~\system{} offers a practical path toward socially deployable proactive conversational agents.

\end{abstract}

\section{Introduction}
Artificial Intelligence (AI) has voice assistants evolving from reactive query–response models to proactive support that anticipates intent and resolves ambiguities during live interaction~\cite{chen2025llamapie, ebert2025reading, acikgoz2025proactive}. Always‑listening assistants enable this shift, but raise serious privacy concerns in social settings, where speech capture may inadvertently include non‑consenting participants. Because speech is deeply personal and identity‑revealing, this tension poses a central challenge to deploying proactive assistants in practice.

In this work we present \system, a framework which attempts to answer the following: \textit{How can proactive speech-capture AI assistants anticipate user needs and coordinate follow‑ups while respecting real‑world privacy constraints?}
\system{} aims to unlock such privacy-centric AI assistants through a three-pronged approach. 
% \system first constrains each assistant to, at all times, capture only the speech of the device owner through real-time speaker verification. Simply put: my assistant has permission to always capture only my speech. However, by capturing one-sided speech \system is now faced with missing context during conversations.
% Second, to recover missing context we build context-gap identification and spatio-temporal context resolution mechanisms that detect missing information and either infer it from available context or request it from interlocutors' agents. 
% Third, we enable assistant-to-assistant (A2A) communication through a privacy-aware trust model that governs implicit information sharing, allowing an agent to assess trust levels when responding to information requests from other \system assistants. 
First, each assistant continuously captures only the device owner’s speech via real‑time speaker verification, ensuring strict ownership boundaries. This one‑sided capture, however, introduces missing conversational context. Second, \system detects and resolves these context gaps using spatio‑temporal reasoning, inferring information when possible or issuing targeted queries to interlocutors’ agents. Third, \system enables privacy‑aware assistant‑to‑assistant (A2A) communication governed by a relationship‑based trust model that controls what information may be safely shared.
For example: In a conversation between Bob and Alice, Bob’s \system{} operates in the background, listening only to Bob. Asynchronously, it identifies gaps, resolves local references, and formulates targeted queries to Alice for the missing information.
Alice’s \system checks these queries against relationship-based permissions and responds only with information that can be shared. Through this A2A exchange, Bob's agent can reconstruct missing context, for example: \emph{..meet at..8 AM}, without Alice's audio.

Our approach reframes always-on AI not as a surveillance problem, but as a coordination problem between privacy-preserving assistants, striking a practical balance between continuous intelligence and consent-driven data sharing.

\section{The Concord Architecture}
\begin{figure*}
\centering
\includegraphics[width=\linewidth]{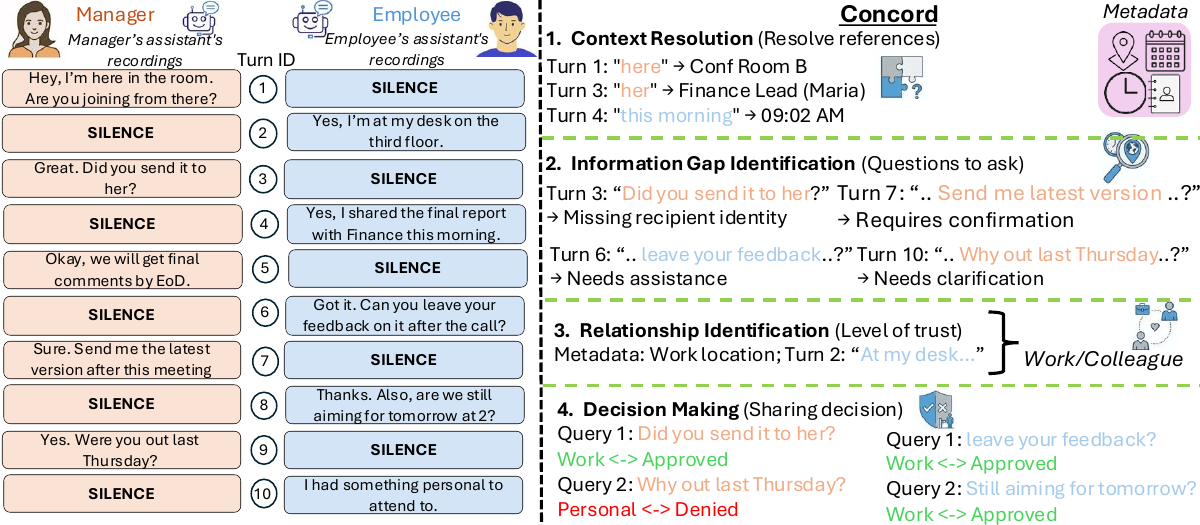}
\caption{\system{}: one-sided transcript processing pipeline. (Left) Multi-turn two-user dialogues, where each user's assistant only records their authenticated user. (Right) The system performs spatio-temporal reference resolution and identifies information gaps (\ref{sec:info_gap}), then applies a decision filter to approve queries based on the relationship (\ref{sec:decision}).}
\label{fig:concept}
\end{figure*}

\system{}'s foundation is privacy-centric real-time speaker verification, which results in a one-sided transcript for the authenticated user (henceforth: user transcript). Following this, the objective of \system{} is to detect information gaps in this transcript, and resolve them through relationship and sensitivity-driven A2A collaboration. Figure~\ref{fig:concept} shows our system pipeline through an example. The following subsections delve into the details of this approach and implementation.
% The assistant must reason over incomplete dialogue and determine when critical context is missing.
% motivating the need for an assistant-to-assistant (A2A) collaboration framework.
% The objective of \system{} is to detect and resolve implicit references in this partial transcript, and to identify information gaps that cannot be filled locally. 
% \system{} addresses these gaps through structured A2A collaboration, allowing missing context to be recovered. 

% This is carried out in two stages. In the first stage, \system analyzes each dialogue turn to detect implicit references (e.g., pronouns, temporal expressions, spatial phrases) and attempts to resolve them using the one-sided script and available mobile metadata, such as current time and location. When a reference cannot be resolved locally, the system flags it as an information gap and queries the interlocutor's assistant.
% In the second stage, when an assistant receives a query requesting information, it evaluates whether the requested information can be disclosed. Based on the relationship between participants, prior interaction context, and the sensitivity of the requested content, the assistant may decide to accept or deny the request. 
\subsection{Speaker Verification }
Speaker verification as well as identification are well-established research areas with extensive prior work~\cite{wang2023wespeaker, liu2023disentangling}. We implement speaker verification using the ECAPA-TDNN architecture~\cite{desplanques2020ecapa} at the voice recording end. We selected the this model for its lightweight and strong performance. It operates as the primary privacy filter for ~\system with a goal to ensure isolation and recording of only the authorized user. For on-device use, we deployed the model on Android (Samsung S20 Ultra) using TFLite optimizations. The system achieves a Real-Time Factor~(RTF) of 0.08, allowing it to run real-time in the background with minimal computational overhead. In our framework, false positives are a risk, as recording an unauthorized speaker may lead to privacy violations. Hence, we prioritize minimizing false positives, even at the cost of a lower true positive rate (TPR). To enforce this constraint, we fix the False Positive Rate~(FPR) at 1\% and select the corresponding decision threshold.

\subsection{Context Resolution and Information Gap Detection} \label{sec:info_gap}

This stage resolves implicit context references and identifies information needed from the interlocutor for completing actionable items. We perform this in two steps, both implemented using few-shot GPT-4.1 with a low temperature ($T=0.1$) to prioritize deterministic outputs~\cite{renze2024effect}.

\noindent $\blacksquare$ \textbf{Local Reference Resolution}:  
This step combines the user transcript with mobile metadata (location, time, calendar) to resolve implicit references and identify information gaps that require input from interlocutor. We first resolve ambiguous linguistic expressions such as pronouns, temporal phrases, and spatial mentions by grounding them in both prior dialogue turns and real-world metadata (including GPS location, Wi-Fi SSID, and calendar events). For example, an utterance such as \emph{I'll go there} links \emph{there} to a specific venue inferred from recent dialogue and current location, while phrases such as \emph{after that} or \emph{tomorrow} are mapped to concrete calendar events when possible. This step determines whether a missing entity can be fully specified using only locally available conversational and mobile data information.

\begin{table*}[ht]
\centering
\small
\setlength{\tabcolsep}{4pt}
\begin{tabular}{>{\centering\arraybackslash}p{1.4cm} p{2.0cm} p{6.5cm} c c c}
\toprule
\textbf{Relation Level} & \textbf{Proximity} & \textbf{Linguistic Indicators} 
& \multicolumn{3}{c}{\textbf{Disclosure Policy by Sensitivity}} \\
\cmidrule(lr){4-6}
 &  &  & \textbf{Low} & \textbf{Mid} & \textbf{High} \\
\midrule

\rotatebox{90}{$L_1$} & 
Intimate (High) &
Implicit references to private spaces (\emph{the bedroom}); elliptical commands (\emph{grab that}); relational nouns (\emph{mom}, \emph{babe}); high ambiguity tolerance due to shared grounding. &
Direct Reveal & Direct Reveal & Direct Reveal \\

\rotatebox{90}{$L_2$} &
Social (Moderate) &
Balanced shared history; first-name address; conversational but more explicit entity references; moderate formality. &
Direct Reveal & Approval Loop & Suppress \\

\rotatebox{90}{$L_3$} &
Professional (Low) &
Formal register; honorifics (\textit{Dr., Mr., Ms.}); task-oriented vocabulary; distancing modal verbs (\textit{could you, would you}). &
Approval Loop & Suppress & Suppress \\

\bottomrule
\end{tabular}
\caption{Unified framework: relationship level inference (left) and sensitivity-aware disclosure policy (right).}
\label{tab:unified_policy}
\end{table*}

\noindent $\blacksquare$ \textbf{Information Gap Detection}: After attempting local resolution, \system{} identifies entities that remain unresolved from the user transcript and available metadata. We perform turn-level analysis using a sliding window over the current turn and the previous five turns. This focuses reasoning for identifying missing context on the most relevant context while limiting interference from distant dialogue.

Rather than asking the model whether a sentence contains sufficient detail to resolve context in a user transcript, we define explicit attribute requirements for each common entity type.
(a) Medical entities require name, dosage, and frequency;  
(b) Temporal entities require event-level specificity beyond relative expressions;  
(c) Spatial entities require building-, floor-, or room-level identifiers;  
(d) General coordination entities require identifying attributes for people, objects, or tasks.

Entities that still fail these requirements are labeled as \emph{information gaps}. The system then queries the interlocutor’s assistant for missing attributes under the disclosure policy described next. At this stage, we prioritize recall to avoid missing critical entities, allowing later stages to handle unnecessary queries.

\subsection{Privacy Management} \label{sec:decision}
% Having identified information gaps, the next stage manages A2A collaboration and the social rules for sharing information.
% % It determines if the information requested can be safely shared within the current social setting. 
% This process comprises of two parts;
% Relationship Assessment: Analyses the one-sided conversation and assigns the interlocutor a trust level.
% Hybrid Relationship-based Disclosure: A decision block that weighs data sensitivity against social expectations to decide if information can be safely shared with interlocutor.
Next, \system manages A2A collaboration to fill these gaps by assessing interlocutor trust level and enforcing social rules for safe information sharing. 
% It first assesses the relationship by analyzing the one‑sided conversation to assign an interlocutor trust level. A hybrid relationship‑based disclosure module then balances data sensitivity against social expectations to determine whether the requested information can be safely shared with the interlocutor.

\noindent $\blacksquare$ \textbf{Relationship Assessment}. We categorize interactions into three trust levels, 
% by analyzing the conversational cues within the user’s transcript. These levels are 
modelled after sociological research on social distance, which organizes human interactions based on degree of intimacy~\cite{guerdelli2023interpersonal}. To distinguish between these levels, we utilize three primary indicators: (i) \textit{Shared reference density}, which measures the ratio of implicit to explicit references to identify a shared history between speakers; (ii) \textit{Conversational formality}, which analyzes sentence structure to detect when formal language is replaced by casual phrasing ; and (iii) \textit{Naming conventions}, which tracks the use of title/honorifics versus endearment/first names as a marker of social distance. Table~\ref{tab:unified_policy} outlines the specific indicators used for each relationship level.

\textit{Trust Level Assignment}: We determine the relationship level by aggregating linguistic markers over a sliding window of the conversation. While our current categories are exploratory, they are intended to grow as the system learns from broader deployment. We design the framework to follow a strict privacy-first logic. To prevent accidental disclosure, if the system detects a threshold of distance indicators, the interaction is immediately locked into Level 3 (Professional). Conversely, we prioritize safety by requiring multiple intimacy indicators before promoting an interaction to Level 1 (Intimate). This ensures that isolated collective terms such as \emph{our} do not trigger high-trust sharing prematurely; these words can appear in non-intimate settings too. These linguistic signals serve as high-confidence indicators of trust that operate in addition to any explicit user-defined permissions.

\noindent $\blacksquare$ \textbf{Hybrid Relationship-Based Disclosure} We use a two‑stage disclosure filter that balances strict privacy guarantees with social flexibility. High‑risk data is protected by deterministic hard locks that override any social reasoning, while lower‑risk interactions incorporate contextual relationship analysis to enable helpful, situation‑appropriate responses. We propose privacy as a dynamic social negotiation rather than a static setting, adapting disclosure to the current context without compromising sensitive information.

\textit{Deterministic Hard Lock:} This filter acts as a high-security safeguard for data with zero tolerance for social disclosure, such as banking credentials or government IDs. We use a deterministic override to prevent the system from considering social norms for these critical items:

{\centering
$\displaystyle
D_{\text{gate}}(E, L) =
\begin{cases}
\text{Abort} & \text{if } \sigma(E) > L \\
\text{Proceed} & \text{otherwise}
\end{cases}
$
\par}

If the requested information is classified as high-sensitivity (Level $L_1$) and the interlocutor is not an intimate contact (assigned to $L_1$), the system immediately terminates the proactive prompt to prevent any risk of exposure.

\textit{Fuzzy Social Norms Matrix:}  Requests that pass the initial security filter are evaluated against a $3 \times 3$ Social Sharing Matrix (Table~\ref{tab:unified_policy} (right)). This matrix maps our three relationship levels ($L_1, L_2, L_3$) against the sensitivity of the information (Low, Mid, High) to select a disclosure protocol. A Privacy Intent Detector provides an additional safeguard: deliberate abstraction, self‑censoring language, or avoidance cues by the user(e.g., “I had some personal stuff”) automatically elevate sensitivity. 
% Deliberate abstraction, self-censoring language, or explicit avoidance cues signal that the user prefers restricted disclosure;it automatically raises the information's sensitivity level. 
This causes the assistant to mirror the user’s caution, shifting from direct responses to permission‑based or no responses. The final decision yields (Table~\ref{tab:unified_policy})) one of four outcomes: (1) Direct Reveal: Full information is shared, (2) Partial Reveal with sensitive entities hidden, (3) Approval Loop: via an explicit private signal (haptic buzz) before speaking. (4) Suppression: silence to protect owner's privacy.

\begin{comment}
\begin{table}
\centering
\small
\setlength{\tabcolsep}{4pt}
\begin{tabular}{lccc}
\toprule
\textbf{Trust} & \textbf{Low} & \textbf{Mid} & \textbf{High} \\
\textbf{Tier}  & \textbf{(Logistics)} & \textbf{(Functional)} & \textbf{(Private)} \\
\midrule
$L_1$ &
Direct Reveal &
Direct Reveal &
Direct Reveal \\

$L_2$ &
Direct Reveal &
Approval Loop &
Suppress \\

$L_3$ &
Approval Loop &
Suppress &
Suppress \\
\bottomrule
\end{tabular}
\caption{Decision policy based on information sensitivity and relationship tier.\ts{combine table 1 and 2?}}
\label{tab:decision_policy}
\end{table}
\end{comment}
\section{Dataset and Baselines}\label{sec:data}

\begin{table}[ht]
\centering
\scriptsize 
\setlength{\tabcolsep}{2pt} 
\renewcommand{\arraystretch}{1.3} 
\begin{tabularx}{\columnwidth}{l l c X c c}
\toprule
& \textbf{Interaction} & \makecell{\textbf{Turns} \\ \tiny Med (Std)} & \textbf{Topics} & \makecell{\textbf{Dur. (min)} \\ \tiny Med (Std)} & \makecell{\textbf{Act.} \\ \tiny (\%)} \\
\midrule

% Using [t] to align with the top row (Dr.-patient)
\multirow[t]{2}{*}{\textbf{High-st.}} 
& Dr.-patient & 51.0 (8.9) & Headaches, Diabetes, Surgery & 15.3 (2.7) & 69.3 \\
& Law.-client & 56.0 (8.7) & Contract, Payment, Property & 16.8 (2.6) & 78.3 \\
\midrule

\multirow[t]{3}{*}{\textbf{Work}} 
& Colleague & 55.0 (8.5) & Tasks, Exp., Manuscript & 16.5 (2.5) & 82.4 \\
& Client-dev & 52.5 (2.7) & Dashboard, Bugs, UAT & 15.8 (0.8) & 58.3 \\
& Mgr.-emp. & 48.5 (2.5) & Deliverables, Budget, Clients & 14.6 (0.8) & 60.0 \\
\midrule

\multirow[t]{2}{*}{\textbf{Academic}} 
& Teach-stud & 49.5 (17.3) & Density, Evolution, Circuits & 14.9 (5.2) & 52.6 \\
& TA-student & 50.0 (6.3) & Chem., Research, Prog. & 15.0 (1.9) & 66.7 \\
\midrule

\multirow[t]{2}{*}{\textbf{Informal}} 
& Friends & 75.5 (12.6) & Movies, Vacation, Games & 22.7 (3.8) & 36.4 \\
& Housemate & 56.5 (14.9) & Cleaning, Bills, Groceries & 17.0 (4.5) & 45.6 \\
\bottomrule
\end{tabularx}
\caption{Dataset statistics for the~\system{} framework. High-st.: High-stakes and Act.: Actionable Items.}
\label{tab:conversation_stats}
\end{table}

\subsection{Dataset}
~\system{} requires two types of datasets: (i) a speaker verification dataset, and (ii) a turn-based dialogue dataset involving conversations between two users. We use speech between two interlocutors as a preliminary exploration case.

\noindent \textbf{Dataset for speech verification:} We use the VoxConverse~\cite{chung2020spot} dataset, which comprises 50 hours of multi-speaker, turn-based conversational audio extracted from YouTube videos. We employ the ECAPA-TDNN model pre-trained on the VoxCeleb~\cite{nagrani2017voxceleb} dataset for target speaker verification. 

\noindent \textbf{Dialogue dataset generation:}
Existing dialogue datasets~\cite{li2017dailydialog} are not suited for studying context resolution, information gap detection, relationship-aware reasoning and decision-making. While several synthetic and semi-synthetic resources exist~\cite{wu2024autogen, chen2025llamapie, ge2025tremu}, they are optimized for response generation or task completion and lack explicit, fine-grained annotations for conversational resolution and information gaps. In particular, they do not encode missing temporal cues, spatial references, contextual dependencies, implicit assumptions about relationships. As a result, they do not provide the ground truth necessary to evaluate structured gap detection and relationship-based decision policies.

To address this limitation, we construct a structured synthetic dataset with explicit control over contextual dependencies and information gaps. Rather than relying on post-hoc LLM annotation, we design an event-based generation pipeline in which the model first produces structured answers and gap specifications, and later generates dialogues conditioned on these constraints. This reduces hallucinations and ensures faithful realization of intended resolution requirements.
The resulting dataset contains approximately 5,700 dialogues spanning nine relationship types across high-stakes, workplace, academic, and informal settings (Table~\ref{tab:conversation_stats}). We validate annotation quality through evaluation against 500 sampled dialogues independently labelled by three students, achieving a Cohen’s Kappa score of 0.78 (good agreement)~\cite{li2024generation}.

\subsection{Baselines}

We evaluate \system{} using three widely adopted reasoning-based resolution techniques: \textit{Chain-of-Thought} (CoT)~\cite{wei2022chain}, \textit{Chain-of-Thought with Self-Reflexion} (CoT-SR)~\cite{shinn2023reflexion}, and \textit{Tree-of-Thought} (ToT)~\cite{yao2023tree}. 
\textbf{CoT }%Recent work on mathematical problem solving and question-answering tasks~\cite{wei2022chain, zhouleast} shows that CoT improves performance compared to standard LLM prompting by encouraging the model to reason through intermediate steps while answering a query. 
%We use CoT to evaluate whether step-by-step reasoning reduces the probability of hallucination and enhances the model’s ability to detect missing or ambiguous information in conversations.
allows the LLM to generate intermediate reasoning steps before producing the final answer.
\textbf{CoT-SR} method extends CoT performance by enabling the model to evaluate and refine its reasoning through self-reflection. 
%We use it to evaluate whether iterative reasoning refinement improves the performance further in information gap resolution.
\textbf{ToT} method allows the model to evaluate alternative reasoning paths before selecting a final output~(see $\S$\ref{sec:reason_base} for details).

%In addition to these reasoning-based approaches, 

We further evaluate \system{} on other LLM variants- GPT-4o, Claude-3.5 Sonnet, and Gemini-1.5. We report the performance of each LLM alongside its best-performing reasoning-based technique~(shown in Table~\ref{table:full_results}). %While several variants of CoT~\cite{zhangautomatic, lyu2023faithful, wangself}, ToT~\cite{mo2024tree} and other reasoning-based techniques~\cite{besta2024graph, chenprogram} exist, 
We limit our evaluation to these variants, compared to~\cite{zhangautomatic, lyu2023faithful, mo2024tree, chenprogram} due to their ease of generalization and consistent performance in diverse conversational scenarios.

\section{Evaluation}
\begin{table*}
\centering
\small
\begin{tabularx}{\textwidth}{@{}llCCCC@{}}
\toprule
\textbf{Framework Component} & \textbf{Evaluation Metric} & \textbf{Zero-shot LLM} & \textbf{SOTA~(CoT-SR)} & \textbf{~\system{}} \\
\midrule
\multirow{2}{*}{Local Reference Resolution} 
 & True Positive Rate (TPR \% $\uparrow$) 
 & 28.4 & 35.49 & \textbf{78.3} \\
 & Sentence Similarity (\% $\uparrow$)    
 & 34.9 & \textbf{65.7} & 58.1 \\
\midrule
\multirow{2}{*}{Info Gap Detection} 
 & True Positive Rate (TPR \% $\uparrow$) 
 & 46.2 & 45.8 & \textbf{91.4} \\
 & False Positive Rate (FPR \% $\downarrow$) 
 & 22.8 & 19.5 & \textbf{6.8} \\
\midrule
Relationship Assessment         
 & Classification Accuracy (\% $\uparrow$) 
 & 95 & 97.0 & \textbf{96.4} \\
\midrule
\multirow{2}{*}{Hybrid Decision Gate} 
 & True Negative Rate (TNR \% $\uparrow$) 
 & 66.5 & 94 & \textbf{97.0} \\
 & True Positive Rate (TPR \% $\uparrow$) 
 & 56.3 & 12 & \textbf{86.0} \\
\bottomrule
\end{tabularx}
\caption{System-level performance across the different components of~\system{}. Our approach achieves high Safety Integrity (97.0\% TNR) while maintaining high utility for proactive gap resolution.}
\label{table:full_results}
\end{table*}

We evaluate~\system{}'s four primary components: context resolution, information gap detection, relationship identification, and final disclosure decision (\autoref{table:full_results}). Our evaluation focuses on the~\system{}'s ability to maintain high utility in identifying needs while prioritizing safe information sharing. 

\subsection{Speaker Verification Accuracy}

We evaluate the speaker verification model on the VoxConverse test dataset, which comprises multi-speaker audio extracted from 232 YouTube videos. The dataset spans 2,612 minutes and includes 8,268 annotated speaker turns. The model is applied to an audio signal using a sliding window with a window length of 2 seconds and an overlap of 0.5 seconds (empirically optimized). This configuration results in $\sim$450 evaluated audio segments per recording on average, yielding $\sim$104,400 evaluated segments across the dataset. The model achieves a 0.8\% FPR and a 99.2\% TPR. Our tuning results in a False Negative Rate (FNR) of $\sim$12\%, which we consider an acceptable trade-off. We adopt a conservative recording threshold to minimize the risk of capturing bystander audio, even if it occasionally results in some missed recordings of the user.

%We tested speaker verification implementation over 28,000 samples, achieving a 0.8\% False Positive Rate (FPR) and a 99.2\% True Positive Rate (TPR). While our specific tuning results in a False Negative Rate (FNR) of approximately 12\%, we consider this an acceptable trade-off. We prioritize a conservative recording threshold to ensure that bystander audio is never mistakenly captured, even if it occasionally results in the system failing to record the target user.

\noindent \textbf{Spatio-Temporal Context Resolution:} We first evaluate the system's ability to identify and ground ambiguous references within the user transcript. The resolution engine achieves a 78\% TPR in identifying turns that require context resolution.
For these identified references, the system produces resolutions with a 58\% semantic text similarity compared to the ground truth. While linguistic variation in location descriptions (e.g., \emph{Clinic Hallway} vs. \emph{North Wing Corridor}) limits exact-match performance, the 78\% TPR ensures that the majority of information gaps are flagged before data is passed to the next stage.

\noindent \textbf{Information Gap Detection:} We prioritize recall in gap detection to ensure no critical (e.g. medical or logistical) detail is overlooked. The system achieves a 91.4\% TPR, successfully capturing every information gap defined in our test set. This aggressive detection strategy results in a 6.8\% FPR. We consider this an acceptable trade-off, because a false positive simply results in a query that the interlocutor’s agent can decline or clarify. This ensures the system acts as an \textit{active investigator} rather than a passive recorder.

\noindent \textbf{Implicit Relationship Identification:} The relationship inference block achieves 96\% classification accuracy across the three trust levels. We observed that misclassifications are almost exclusively directed toward more restrictive levels (e.g., classifying a $L_2$ Social contact as a $L_3$ Professional). This \textit{Safety Bias} is a deliberate design choice. If the linguistic markers are ambiguous, the system defaults to a stricter level to prevent accidental disclosure. Since the interlocutor can always provide further context to re-establish trust, we prioritize this conservative boundary.

\noindent \textbf{Hybrid Decision Gate:} We measure the system's ability to execute the correct disclosure protocol based on the resolved context and inferred relationship. \textit{Utility}: \system{} achieved an 86\% TPR, meaning it correctly identified and shared information in 86\% of cases where disclosure was socially appropriate. \textit{Safety}: The system achieved a 97\% TNR. This high TNR is critical for the \emph{Logic of Privacy}; it demonstrates that in 97\% of sensitive scenarios, the system correctly suppressed or masked information that should not have been shared. These results confirm that the deterministic hard-locks and the fuzzy social matrix effectively guard against privacy breaches in ambient settings.

\section{Related Work}

With Large Language Models (LLMs), assistant-centric frameworks have started to evolve and interact with our daily lives. For example, \textsc{LlamaPIE}~\cite{chen2025llamapie}, DailyLLM~\cite{tian2025dailyllm}, show how agents can be useful for conversation guidance, or automated daily journaling. However, current agent-based systems often assume the availability of reliable and complete user information. In practice, obtaining such contextual data is challenging due to privacy constraints that limit data capture. 

\noindent \textbf{Limitations of Current Access Protocols (MCP).}
%The adoption of large language models (LLMs) has enabled a broad range of real-world applications. 
LLMs have shown potential to support daily-life activities through automated logging of personal information and personalized assistance~\cite{tian2025dailyllm, li2024personal, xu2024autolife}. Model Context Protocol (MCP)~\cite{hou2025model} enable this further by interfacing AI systems with external tools, APIs and sensor-driven resources. Prior efforts~\cite{guo2025sensormcp, yang2025iot} show how MCP-integrated LLMs working with IoT devices and contextual data streams can enable automation. However, these access protocols primarily emphasize connectivity and tool invocation. They do not explicitly model interpersonal relationships, interaction stakes, or socially grounded access constraints. As a result, decisions about information retrieval and sharing are typically governed by static permissions rather than dynamic reasoning.

\noindent \textbf{One-Sided Conversation Reconstruction: Risks of Inference-Based Context Recovery.}
Recent studies on contextual and temporal reasoning~\cite{liubridging, ruiz2025temporal, wang2024tram, ge2025tremu, ebert2025reading} show that LLMs can infer missing or implicit information in conversational settings. Existing approaches often rely on long conversational histories~\cite{liubridging}, persistent user memory~\cite{ge2025tremu}, or stored personal data to fill contextual gaps~\cite{chen2025llamapie}. While these systems improve contextual continuity, they implicitly assume that missing information can be safely inferred from available signals. In practice, however, inference-based reconstruction may lead to incorrect assumptions (e.g., guessing the answer to “What time is the meeting?” without verified access to scheduling details). Existing approaches thus lack mechanisms to request information that requires explicit retrieval.

\noindent \textbf{Gaps of Existing Agent Frameworks.}
Current systems fall short for A2A collaboration: they either retrieve information without considering relations or infer missing details without guarantees of correctness.~\system{} allows for collaborative information access and models human relationships and the constraints they impose.

 \section{Discussion
 and Conclusion}
 \system is a framework for socially deployable proactive AI assistants built on one‑sided speech capture, information‑gap identification, context resolution, and A2A communication. Our exploratory evaluation highlights the capabilities enabled by this design and its current limitations, motivating several directions for future work:

\noindent \textbf{Multi-speaker extensions:} Evaluating \system in group conversations introduces challenges in isolating the primary user’s voice under overlapping speech, tracking turn-taking dynamics, and attributing references to speakers.
% and correctly identifying interlocutors for asynchronous follow‑up.
% The evaluation of \system in a group conversation will push the boundary on effectively capturing just the primary user's voice. Further, it brings up the need to correctly identify other interlocutors for asynchronous follow-up.

\noindent \textbf{Cross‑modal context recovery:} Beyond spatial and temporal cues, integrating modalities such as vision and digital context (email, text, calendar) can improve priority detection, task identification, and context resolution.

\noindent \textbf{Interlocutor agent discovery and connection:} Reliable participant identification and asynchronous agent connectivity are critical, particularly in multi‑speaker settings where dynamic participation complicates A2A coordination.
% research around connectivity and device discovery to reliably identify participants in a conversation, and carry out asynchronous connection with their agents to resolve context gaps. This becomes an interesting challenge particularly in the multi-speaker conversation setting.

\noindent \textbf{Evolving minimal-disclosure protocols:} \system{} uses intimacy level to gate information-disclosure (largely driven by boundaries established through digital context sharing research). Socially deployed always‑listening assistants demand a deeper understanding of social consequences, trust tiers, and the broader privacy–convenience trade‑off.

\noindent \textbf{Longitudinal cross-conversation context building:} Proactive assistants should accumulate short‑ and long‑term context over time, learning user preferences from speech and multimodal signals to enable personalization.
% The ultimate intent of proactive AI like \system is to always be available, listen to the user, automate everyday tasks, and build context over time for personalization. This involves both short-term and long term context building. Evolving versions of \system will target learning on the go where the system learns user preferences from their speech, and articulation, connect with other modalities to personalize the AI assistant.

% 1. multi-speaker and group conversation extension
% 2. Richer information gap modelling, cross-modality context recovery with multi-point verification for better accuracy
% 3. Connection and discovery model for interlocutor agents
% 4. Developing standardized minimal-disclosure protocols. Deeper understanding of trust tiers and information boundaries
% % 7. Governance and accountability for proactive AI assistants
% 5. 
By framing missing context as information gaps and enforcing relationship-aware disclosure policies,~\system{} demonstrates that proactivity, privacy, and usability can coexist.

\appendix

\section{Dataset generation}
\label{sec:appendix}

\subsection{Keywords generation}
For generating synthetic key words for a target conversation, we use the publicly available dataset for that relationship type. Some of the public available dataset are- DailyDialogy, ChatDoctor.
We analyse the dialogue dataset and derive contextually relevant keywords

\subsubsection{Keywords generation prompt} 

\begin{itemize}[noitemsep,leftmargin=*,topsep=0pt]
    \item \textbf{Keywords generation from publicly available datasets for teacher--student interaction}
    \begin{enumerate}

\item Analyse the conversation between the teacher and the student.

\item Generate keywords based on the content of the conversation.

\item Extract a list of educationally and conversationally relevant keywords or short phrases from the conversation. These keywords may include learning topics, academic concerns, study strategies, assignments, feedback, classroom activities, performance discussions, or instructional guidance provided by the teacher. Generic or filler words (e.g., ``the,'' ``and,'' or ``okay'') must be excluded.

\item For each extracted keyword, assign a score from 1 to 10 indicating how commonly the keyword appears in real-world teacher--student conversations, where:
\begin{itemize}
    \item 1 = Rarely used in teacher-student academic discussions
    \item 10 = Very frequently used and standard in classroom or academic conversations
\end{itemize}

The assigned ranking should reflect how commonly the keyword appears in typical educational interactions rather than how often it appears in the specific conversation.

\item Do not assume or infer academic problems or learning issues beyond what is explicitly stated or strongly implied in the conversation.

\item Use neutral, professional, and education-appropriate language.

\item Arrange the extracted keywords in descending order based on their assigned rank such that:
\begin{itemize}
    \item Keywords with a rank of 10 appear first
    \item Keywords with lower ranks follow in decreasing order (9 $\rightarrow$ 1)
    \item Keywords with the same rank may appear in any order relative to each other
\end{itemize}

\item Generate exactly 500 keywords.
\end{enumerate}

\item \textbf{Keywords generation with explicit prompt for housemates interaction}

\begin{enumerate}

\item Consider a conversation occurring between companions such as couples, flatmates, siblings, or individuals sharing a close personal or living relationship.

\item Generate keywords based on the semantic and conversational content of the dialogue.

\item Extract a list of lifestyle-focused, relationship-oriented, household coordination, emotional support, and daily life interaction keywords or short phrases from the conversation. Keywords may include, but are not limited to:
\begin{itemize}
    \item Household responsibilities or shared chores
    \item Financial or expense planning discussions
    \item Daily routine coordination or schedule planning
    \item Emotional support or wellbeing discussions
    \item Relationship communication or personal interaction topics
    \item Living arrangement coordination or household management
    \item Social or family event planning
    \item Personal habits or lifestyle adjustment discussions
    \item Shared resource or space usage coordination
    \item Decision-making related to travel, purchases, or activities
    \item Conflict resolution or compromise discussions
    \item Personal health, wellness, or daily care coordination
    \item Communication, trust, or relationship maintenance topics
\end{itemize}

Generic or filler words (e.g., ``the,'' ``and,'' ``okay,'' ``yes'') must be excluded.

\item For each extracted keyword, assign a score from 1 to 10 indicating how commonly the keyword appears in real-world companion or close personal relationship conversations, where:
\begin{itemize}
    \item 1 = Rarely used in everyday companion or shared living conversations
    \item 10 = Very frequently used and standard terminology in household or close relationship interactions
\end{itemize}

The assigned ranking should reflect typical conversational frequency in companion relationships rather than how often the keyword appears within the specific dialogue.

\item Do not infer or fabricate relationship conflicts, emotional distress, or personal issues beyond what is explicitly stated or clearly implied in the conversation.

\item Use neutral, respectful, and socially appropriate conversational language.

\item Arrange the extracted keywords in descending order based on their assigned rank such that:
\begin{itemize}
    \item Keywords with a rank of 10 appear first
    \item Keywords with lower ranks follow in decreasing order (9 $\rightarrow$ 1)
    \item Keywords with the same rank may appear in any order relative to each other
\end{itemize}

\item Generate exactly 500 keywords or short phrases.

\textit{Note:} The number of keywords is fixed at 500 to minimize the likelihood of hallucinated outputs and the inclusion of non-essential or irrelevant terms.

\end{enumerate}
\end{itemize}

\subsubsection{Example keywords}
\begin{itemize} [noitemsep,leftmargin=*,topsep=0pt]
    \item \textbf{Sample keywords generated after analysing a publicly available dataset~(Teacher--student~\cite{shani2405multi})}
    
    Discussion, experiment, lecture, assessment, activities, timeline, question, answer, group work, interactive, hands-on, assignment, feedback, quiz, summary, example, role-play, classroom, lesson, research, presentation, worksheet, class, student, teacher, understanding, concept, topic, learning, instructions, activity, study, accomplishments, challenges, creative, motivation, problem, solution, performance, review, application, context, questions, history, science, art, method, group, strategy, critical thinking, politics, democracy, freedom, revolution, war, government, civil rights, independence, conflict, equality, leader, values, beliefs, social, economic, oppression, rights, historical, engineering, architecture, society, environment.

    \item \textbf{Sample keywords generated by explicit prompt to an LLM~(Housemates)}
    
    Chores, cleaning, cooking, groceries, dishes, laundry, bills, rent, budget, dinner plans, schedule, bedtime, wake-up time, work hours, shopping list, trash, takeout, breakfast, lunch, dinner, pets, errands, cleaning schedule, shared expenses, grocery shopping, household supplies, utilities, Wi-Fi, dishes rotation, bathroom cleaning, vacuuming, recycling, laundry day, meal prep, food preferences, cooking rotation, shared meals, grocery run, fridge space, food storage, cleaning products, rent
    
\end{itemize}

\subsection{Event generation}
We utilize the generated keywords to generate fictional events related to that relationship.

\subsubsection{Event generation prompt}
\begin{itemize}

\item Analyse the provided keywords and construct potential user profiles.

\item Generate ten events based on the keywords, modelling interactions between a teacher and a student.

\item Carefully analyse each keyword and curate suitable teacher and student profiles such that their roles, academic goals, and learning challenges logically align with the provided keywords.

\item Ensure that each generated event reflects realistic classroom environments, academic expectations, mentoring relationships, and authentic learning experiences.

\item Output the events in the following structured JSON format:
\begin{verbatim}
{
    { ... },
    { ... },
    ...
    { ... }
}
\end{verbatim}

\item Each event must be written as a narrative story.

\item Each event must begin with the following profile paragraph, followed by a continuous narrative:

\textit{[Name] is a [age]-year-old [occupation] living in [location].}

\textbf{Story Style Guidelines}

\begin{itemize}

\item Write the narrative in third-person perspective.

\item The story should naturally integrate:
\begin{itemize}
    \item Student thoughts, motivations, learning concerns, and academic pressures
    \item Teacher expectations, mentoring approaches, and instructional strategies
    \item Classroom dynamics, peer collaboration, and educational environment
\end{itemize}

\item Avoid including:
\begin{itemize}
    \item Explicit academic challenges or learning conflicts
    \item Feedback, grading, or performance evaluation discussions
    \item Instructional or academic decision-making processes
    \item Student growth, skill development, or resolution of challenges
    \item Classroom-, institutional-, or peer-level outcome summaries
\end{itemize}

\item Additional formatting constraints:
\begin{itemize}
    \item Do not use dialogue labels such as ``Teacher:'' or ``Student:''
    \item Do not use bullet points or section headings within the narrative
    \item Avoid formal evaluation or administrative report writing style
    \item Maintain a natural, coherent storytelling format
\end{itemize}

\end{itemize}

\item Each generated event must contain a minimum of 500 words.

\end{itemize}

\subsubsection{Example event}
Evelyn is a 14-year-old student living in Boston. The classroom was alive with energy as the students gathered for the hands-on science lesson. Today, Evelyn’s teacher, Ms. Garcia, had carefully prepared an experiment to help her students better understand the concept of buoyant force and density. Evelyn watched as Ms. Garcia demonstrated Archimedes' Principle using a tub of water, various objects, and a spring scale. The teacher’s approach was interactive, inviting students to predict which objects would float and which would sink before testing their hypotheses. Evelyn felt a surge of excitement; she had always been curious about the science behind everyday phenomena. As her group worked through the experiment, she collaborated with her peers to record displacement and calculate weight, eager to see if their predictions matched the results. The teacher moved from group to group, observing their progress and offering encouragement, fostering a supportive environment where curiosity was celebrated. Evelyn felt motivated by the teamwork and the practical nature of the activity. The lesson ended with students sharing their findings, and Evelyn reflected on how the experiment deepened her understanding of the scientific method. She left class looking forward to more opportunities for discovery and hands-on learning.

\subsection{Dialogue generation}
\subsubsection*{Core Logic and Constraints}

\paragraph{1. The Core Logic (Read Carefully)}
\begin{itemize}
    \item \textbf{The ``Gap'':} Assistant A records \textbf{only} User A. Assistant B records \textbf{only} User B. Each assistant cannot hear the other user.
    
    \item \textbf{The Problem:} When User B says ``Let's meet at \textit{Joe's Pizza},'' Assistant A hears silence, followed by User A saying ``Okay.'' Assistant A therefore misses the location entity.
    
    \item \textbf{The Goal:} Generate the full conversation along with the specific \textbf{Protocol Queries} required to resolve the missing contextual gaps.
\end{itemize}

\paragraph{2. Strict Constraints}
\begin{itemize}

\item \textbf{Length:} Approximately 2,000 words (approximately 15 minutes of dialogue).

\item \textbf{High Density Requirement:}
Include at least 15 local context resolutions and 8 inter-agent queries.

\textbf{Required Resolution Types (Critical):}
\begin{itemize}
    \item \textbf{Spatial} references (e.g., here, there, this place, that building, the café, that intersection, etc.)
    \item \textbf{Temporal} references (e.g., now, then, later, today, tomorrow, at that time, etc.)
\end{itemize}

\item \textbf{Query Quality Filter (Critical for Training):}
Generate two types of potential protocol queries:

\begin{itemize}
    \item \textbf{HIGH\_VALUE Queries} (8+ queries):  
    Queries that involve missing information that is either:
    \begin{itemize}
        \item Actionable (e.g., time, location, contact name)
        \item Sensitive (e.g., privacy-related information)
    \end{itemize}

    \item \textbf{LOW\_VALUE Queries} (3--5 queries):  
    Queries that involve missing information that is not important, such as:
    \begin{itemize}
        \item Jokes
        \item Weather comments
        \item Casual observations
        \item Opinions or small talk
    \end{itemize}

    \item Each generated query must include the field:
\begin{verbatim}
query_quality_check: "HIGH_VALUE"
\end{verbatim}
or
\begin{verbatim}
query_quality_check: "LOW_VALUE"
\end{verbatim}

\item \textbf{Purpose:}  
The LOW\_VALUE queries are used to evaluate whether the Information Gap Detection system can correctly reject irrelevant or non-actionable queries.

\end{itemize}

\end{itemize}
\subsubsection{Generation steps}
\textbf{Step 1: The World State} \\
Define a complex scenario (e.g., organizing a protest, planning a secret surprise, managing a crisis).\\ 
\textbf{Step 2: The Metadata} \\
Generate valid JSON metadata for \textbf{both} users (e.g., GPS, Wi-Fi, calendar) that explains the local context. For example, specify why a spatial reference such as ``here'' corresponds to a specific location (e.g., ``Central Park'').\vspace{0.5em}\\
\textbf{Step 3: The Conversation} \\
Write the dialogue while adhering to the following requirements:
\begin{itemize}
    \item \textbf{Force Ambiguity:} Users should rely on implicit context (e.g., ``I'm \textit{here}'', ``Did you ask \textit{him}?'')

    \item \textbf{Force Gaps:} User B should propose plans or introduce names, while User A should agree or respond without explicitly repeating the missing entity.

    \item \textbf{Spatial References (Required):} Include expressions such as ``here'', ``there'', ``this place'', ``that building'', ``the gym'', and ``that corner''.

    \item \textbf{Temporal References (Required):} Include expressions such as ``now'', ``then'', ``later'', ``at that time'', and ``tomorrow morning''.
    \item \textbf{Person References (Required):} Include expressions such as ``him'', ``her'', ``them'', ``that person'', and ``the coordinator''.

    \item \textbf{Object References (Required):} Include expressions such as ``this'', ``that'', ``it'', ``the package'', and ``that document''.
\end{itemize}

\textbf{Step 4: The Answer Keys (Protocol Outputs)} 
\begin{itemize}
    \item \textbf{Resolutions:} Map ambiguous references to explicit contextual entities (e.g., map ``here'' $\rightarrow$ GPS coordinates).

    \item \textbf{Queries:} Generate structured protocol payloads for retrieving missing contextual information.
\end{itemize}

\subsubsection{Output Format}

\begin{tcolorbox}[
  colback=gray!8,
  colframe=gray!40,
  boxrule=0.3pt,
  arc=2pt,
  width=\linewidth,
  enhanced,
  breakable,
  left=8pt,right=8pt,top=8pt,bottom=8pt
]

{\ttfamily\footnotesize
\renewcommand{\arraystretch}{1.45}
\begin{tabular}{@{}p{\linewidth}@{}}
\{
"dataset\_id": "scenario\_protocol\_***",

\vspace{2pt}

"backstory": \{

  "summary": "...",

  \vspace{3pt}

  "relationship": "Colleague | Spouse | Doctor"

\},

\vspace{8pt}

"mobile\_context\_snapshot": \{

  "user\_a": \{

    "location\_semantic": "...",

    \vspace{2pt}

    "gps\_coords": "...",

    \vspace{2pt}

    "wifi\_ssid": "...",

    \vspace{2pt}

    "calendar\_next": "..."

  \},

  \vspace{6pt}

  "user\_b": \{

    "location\_semantic": "...",

    \vspace{2pt}

    "wifi\_ssid": "..."

  \}

\},

\vspace{10pt}

"conversation\_transcript": [

  \{
    "turn\_id": 1,

    "speaker": "User A",

    "text": "..."
  \},

  \vspace{4pt}

  \{
    "turn\_id": 2,

    "speaker": "User B",

    "text": "..."
  \}

],

\vspace{10pt}

"ground\_truth\_resolutions": [

  \{
    "trigger\_turn\_id": 1,

    \vspace{2pt}

    "ambiguous\_phrase": "...",

    \vspace{2pt}

    "resolved\_entity": "...",

    \vspace{3pt}

    "resolution\_source": 
      "User A GPS | User A Wifi | User A Calendar"
  \}

],

\vspace{10pt}

"required\_protocol\_queries": [

  \{

    "trigger\_turn\_id": 3,

    "query\_quality\_check": "LOW | MEDIUM | HIGH",

    "reason": "...",

    "protocol\_payload": \{

      "intent": "...",

      "target\_slot / target\_attribute": "...",

      "urgency": "NONE | ROUTINE | IMMEDIATE"

    \},

    "natural\_language\_fallback": "..."

  \}

]
\}

\end{tabular}
}

\end{tcolorbox}

\subsubsection{Synthetic dialogue}

\begin{itemize}
    \item \textbf{Doctor--patient interaction dialogue}

    \begin{itemize}
  \item \textbf{Turn 1 --- User A:} Hi, Dr. Sharma, I'm here for my appointment.
  \item \textbf{Turn 2 --- User B:} Hello, Samantha, please have a seat. How are you feeling now compared to the weekend?
  \item \textbf{Turn 3 --- User A:} I still have a headache, but it's less intense than it was.
  \item \textbf{Turn 4 --- User B:} Did you notice the dizziness again after you were at that trail?
  \item \textbf{Turn 5 --- User A:} Yes, right after I finished, especially when I stood up by the benches.
  \item \textbf{Turn 6 --- User B:} And how about now?
  \item \textbf{Turn 7 --- User A:} Today, the dizziness is mild. I've been working from my studio, so mostly sitting.
  \item \textbf{Turn 8 --- User B:} Were you able to get enough sleep last night?
  \item \textbf{Turn 9 --- User A:} Not really, I woke up around 3 AM and felt the headache pounding.
  \item \textbf{Turn 10 --- User B:} Did anything help relieve it then?
  \item \textbf{Turn 11 --- User A:} I drank some water and tried breathing exercises, but the pain just eased a little.
  \item \textbf{Turn 12 --- User B:} Did you use any medication this time?
  \item \textbf{Turn 13 --- User A:} No, I didn't want to overdo it on painkillers. I kept some Advil in my bag, just in case.
  \item \textbf{Turn 14 --- User B:} Good, moderation is important. Can you point to which side you feel the pain more now?
  \item \textbf{Turn 15 --- User A:} It's mostly behind my right eye, radiating to my temple.
  \item \textbf{Turn 16 --- User B:} When you say 'that pain', does it feel sharp or dull?
  \item \textbf{Turn 17 --- User A:} Kind of a throbbing ache, not really stabbing.
  \item \textbf{Turn 18 --- User B:} Okay. Last time, you said you had light sensitivity. Is that happening now, here in the office?
  \item \textbf{Turn 19 --- User A:} Not as much now, but at the trail and in my studio yesterday, the sunlight was almost unbearable.
  \item \textbf{Turn 20 --- User B:} Can you tell me how long 'it' lasted yesterday after work?
  \item \textbf{Turn 21 --- User A:} The headache lasted until around 8 PM, so several hours after I left.
  \item \textbf{Turn 22 --- User B:} Did you stay hydrated throughout your ride and afterward?
  \item \textbf{Turn 23 --- User A:} I tried, but I might've slacked a bit. I finished one bottle on the trail, and lost the second in the parking lot.
  \item \textbf{Turn 24 --- User B:} Was that the usual route you took or something new?
  \item \textbf{Turn 25 --- User A:} It was a new loop---I went through Barton Creek, took the creek crossing, and ended up by the old railroad bridge.
  \item \textbf{Turn 26 --- User B:} Were there any moments when you felt you needed to stop or rest more than usual?
  \item \textbf{Turn 27 --- User A:} Yes, especially at that shady spot beneath the large oak. I took five minutes then.
  \item \textbf{Turn 28 --- User B:} And when you noticed 'it' starting, was it before or after you met up with your biking group?
  \item \textbf{Turn 29 --- User A:} After. I joined them late at the far trailhead.
  \item \textbf{Turn 30 --- User B:} Have you had any other episodes like this, say, at work or during a normal day?
  \item \textbf{Turn 31 --- User A:} Yes, once last Thursday, when I was finishing a large project. The headache came on suddenly.
  \item \textbf{Turn 32 --- User B:} Was it similar in intensity to what you felt on the trail?
  \item \textbf{Turn 33 --- User A:} Less intense at work, but still distracting. I had to take a break and step outside.
  \item \textbf{Turn 34 --- User B:} Who was around at the office when that happened?
  \item \textbf{Turn 35 --- User A:} Just my coworker Alex. She offered me some tea.
  \item \textbf{Turn 36 --- User B:} And that seemed to help?
  \item \textbf{Turn 37 --- User A:} Only a little, but I appreciated her help.
  \item \textbf{Turn 38 --- User B:} Earlier, you mentioned 'that folder' had stress-related items. Can you bring it next time?
  \item \textbf{Turn 39 --- User A:} Of course, it’s in my laptop bag.
  \item \textbf{Turn 40 --- User B:} Let’s talk about the testing. Your blood test is scheduled at the lab today, correct?
  \item \textbf{Turn 41 --- User A:} Yes, I’ll head there after this.
  \item \textbf{Turn 42 --- User B:} Was the appointment for 2:00 PM or later?
  \item \textbf{Turn 43 --- User A:} 2:00 PM at Texas Health Lab.
  \item \textbf{Turn 44 --- User B:} And for the MRI, you have a slot at that center I recommended, right?
  \item \textbf{Turn 45 --- User A:} Yes, it’s booked for tomorrow, 4:30 PM at Medical Imaging Center.
  \item \textbf{Turn 46 --- User B:} Good. After the results come in, let’s touch base here or over the phone.
  \item \textbf{Turn 47 --- User A:} Sounds great, I’ll call as soon as I get them.
  \item \textbf{Turn 48 --- User B:} If you notice 'it' worsening, or if there's any new symptom, contact me immediately.
  \item \textbf{Turn 49 --- User A:} Absolutely. Thanks for listening and explaining everything.
  \item \textbf{Turn 50 --- User B:} Of course. By the way, which bike did you use for the ride?
  \item \textbf{Turn 51 --- User A:} My teal Trek. The old one’s still in the shop.
  \item \textbf{Turn 52 --- User B:} I see. Is 'it' heavier than the last?
  \item \textbf{Turn 53 --- User A:} About the same, but it’s got a new hydration system built in.
  \item \textbf{Turn 54 --- User B:} That should help. How long did you ride on that last loop?
  \item \textbf{Turn 55 --- User A:} Almost two hours, with a few stops.
  \item \textbf{Turn 56 --- User B:} Did the terrain seem tougher than usual there?
  \item \textbf{Turn 57 --- User A:} A bit. The creek crossing was muddy, and getting over that bridge took extra energy.
  \item \textbf{Turn 58 --- User B:} Were you riding alone when you felt the worst of it?
  \item \textbf{Turn 59 --- User A:} Yes, for a few miles before the others caught up.
  \item \textbf{Turn 60 --- User B:} Okay. Let's get you ready for the tests. If you need to reschedule, call the front desk or let me know directly.
  \item \textbf{Turn 61 --- User A:} Will do, thanks again.
  \item \textbf{Turn 62 --- User B:} No problem. See you soon, and take care until then.
  \item \textbf{Turn 63 --- User A:} Thank you, Dr. Sharma.
  \item \textbf{Turn 64 --- User B:} You're welcome, Samantha. Enjoy the rest of your day.
  \item \textbf{Turn 65 --- User A:} (smiling) You too. Hopefully the sun won't be so brutal later.
  \item \textbf{Turn 66 --- User B:} (chuckles) Austin weather, can't predict it.
\end{itemize}
\item \textbf{Labelled context resolution}
\begin{itemize}

\item \textbf{Turn 1} --- ambiguous\_phrase: ``here''; resolved\_entity: ``Dr. Sharma's Clinic, West Avenue, Austin''; resolution\_source: ``User B GPS + Wifi''.

\item \textbf{Turn 4} --- ambiguous\_phrase: ``that trail''; resolved\_entity: ``Barton Creek Greenbelt''; resolution\_source: ``User A Calendar''.
\item \textbf{Turn 5} --- ambiguous\_phrase: ``benches''; resolved\_entity: ``Picnic Benches near Creek Crossing, Barton Creek Greenbelt''; resolution\_source: ``User A GPS''.

\item \textbf{Turn 6} --- ambiguous\_phrase: ``now''; resolved\_entity: ``10:30 AM (Time of Appointment)''; resolution\_source: ``User A Calendar''.

\item \textbf{Turn 7} --- ambiguous\_phrase: ``studio''; resolved\_entity: ``Design Studio, East 6th Street, Austin''; resolution\_source: ``User A GPS + Wifi''.

\item \textbf{Turn 9} --- ambiguous\_phrase: ``3 AM''; resolved\_entity: ``03:00 AM, User A's Home, Holly Street''; resolution\_source: ``User A Calendar''.

\item \textbf{Turn 11} --- ambiguous\_phrase: ``water''; resolved\_entity: ``Filtered Water Bottle, 20 oz''; resolution\_source: ``User A Object Log''.

\item \textbf{Turn 13} --- ambiguous\_phrase: ``Advil in my bag''; resolved\_entity: ``Ibuprofen (Advil), 200 mg tablet, Laptop Bag pocket''; resolution\_source: ``User A Object Log''.

\item \textbf{Turn 14} --- ambiguous\_phrase: ``which side''; resolved\_entity: ``Right eye, right temple''; resolution\_source: ``User A Self-report''.

\item \textbf{Turn 15} --- ambiguous\_phrase: ``my right eye''; resolved\_entity: ``Orbital area, right side''; resolution\_source: ``User A Self-report''.

\item \textbf{Turn 16} --- ambiguous\_phrase: ``that pain''; resolved\_entity: ``Headache radiating from right temple, throbbing ache''; resolution\_source: ``User A Symptom Log''.

\item \textbf{Turn 18} --- ambiguous\_phrase: ``here in the office''; resolved\_entity: ``Dr. Sharma's Clinic, Patient Room 2''; resolution\_source: ``User B GPS + Calendar''.

\item \textbf{Turn 19} --- ambiguous\_phrase: ``the trail''; resolved\_entity: ``Barton Creek Greenbelt''; resolution\_source: ``User A Calendar''.

\item \textbf{Turn 19} --- ambiguous\_phrase: ``my studio''; resolved\_entity: ``Design Studio, East 6th Street, Austin''; resolution\_source: ``User A GPS + Wifi''.

\item \textbf{Turn 20} --- ambiguous\_phrase: ``it''; resolved\_entity: ``Headache episode''; resolution\_source: ``User A Symptom Log''.

\item \textbf{Turn 23} --- ambiguous\_phrase: ``one bottle on the trail''; resolved\_entity: ``Hydration bottle, 20 oz, Barton Creek Greenbelt''; resolution\_source: ``User A Object Log + GPS''.

\item \textbf{Turn 23} --- ambiguous\_phrase: ``second in the parking lot''; resolved\_entity: ``Lost water bottle, Parking lot near trailhead''; resolution\_source: ``User A GPS''.

\item \textbf{Turn 25} --- ambiguous\_phrase: ``new loop''; resolved\_entity: ``Barton Creek loop, creek crossing, railroad bridge''; resolution\_source: ``User A Calendar + GPS''.

\item \textbf{Turn 27} --- ambiguous\_phrase: ``shady spot beneath the large oak''; resolved\_entity: ``Large oak tree rest area, Barton Creek Greenbelt''; resolution\_source: ``User A GPS''.

\item \textbf{Turn 28} --- ambiguous\_phrase: ``your biking group''; resolved\_entity: ``Samantha's biking friends: Alex, Jamie, Lee''; resolution\_source: ``User A Calendar''.

\item \textbf{Turn 29} --- ambiguous\_phrase: ``far trailhead''; resolved\_entity: ``South Entrance, Barton Creek Greenbelt''; resolution\_source: ``User A GPS''.

\item \textbf{Turn 34} --- ambiguous\_phrase: ``office''; resolved\_entity: ``Design Studio, East 6th Street, Austin''; resolution\_source: ``User A GPS + Wifi''.

\item \textbf{Turn 35} --- ambiguous\_phrase: ``my coworker Alex''; resolved\_entity: ``Alex N., Graphic Designer, Samantha's colleague''; resolution\_source: ``User A Calendar''.

\item \textbf{Turn 38} --- ambiguous\_phrase: ``that folder''; resolved\_entity: ``Stress Tracking Folder, in Laptop Bag''; resolution\_source: ``User A Object Log''.

\item \textbf{Turn 39} --- ambiguous\_phrase: ``my laptop bag''; resolved\_entity: ``Samantha's black laptop bag (main compartment)''; resolution\_source: ``User A Object Log''.

\item \textbf{Turn 40} --- ambiguous\_phrase: ``lab today''; resolved\_entity: ``Texas Health Lab, 2:00 PM, June 12, 2024''; resolution\_source: ``User A Calendar''.

\item \textbf{Turn 42} --- ambiguous\_phrase: ``appointment''; resolved\_entity: ``Blood Test, Texas Health Lab, 2:00 PM''; resolution\_source: ``User A Calendar''.

\item \textbf{Turn 44} --- ambiguous\_phrase: ``that center I recommended''; resolved\_entity: ``Medical Imaging Center, West Campus, Austin''; resolution\_source: ``User B Calendar''.

\item \textbf{Turn 45} --- ambiguous\_phrase: ``tomorrow, 4:30 PM''; resolved\_entity: ``June 13, 2024, 4:30 PM''; resolution\_source: ``User A Calendar''.

\item \textbf{Turn 46} --- ambiguous\_phrase: ``here''; resolved\_entity: ``Dr. Sharma's Clinic, West Avenue, Austin''; resolution\_source: ``User B GPS + Wifi''.

\item \textbf{Turn 51} --- ambiguous\_phrase: ``My teal Trek''; resolved\_entity: ``Trek bicycle, teal color, used for ride''; resolution\_source: ``User A Object Log''.

\item \textbf{Turn 52} --- ambiguous\_phrase: ``it''; resolved\_entity: ``Trek bicycle''; resolution\_source: ``User A Object Log''.

\item \textbf{Turn 53} --- ambiguous\_phrase: ``new hydration system''; resolved\_entity: ``Integrated hydration reservoir in Trek bike''; resolution\_source: ``User A Object Log''.

\item \textbf{Turn 56} --- ambiguous\_phrase: ``there''; resolved\_entity: ``Barton Creek, creek crossing area''; resolution\_source: ``User A GPS''.

\item \textbf{Turn 57} --- ambiguous\_phrase: ``that bridge''; resolved\_entity: ``Old railroad bridge, Barton Creek Greenbelt''; resolution\_source: ``User A GPS''.

\end{itemize}
\item \textbf{Queries}
\begin{itemize}

\item \textbf{Turn 5} --- query\_quality\_check: ``HIGH\_VALUE''; reason: ``User A referenced the location ('benches') where dizziness was notable, but the specific location is missing for context.''; 
protocol\_payload: intent: ``RESOLVE\_MISSING\_ENTITY'', target\_slot: ``SYMPTOM\_LOCATION'', urgency: ``IMMEDIATE'', context\_ref: ``Turn 5''; 
natural\_language\_fallback: ``Requesting precise location for dizziness episode mentioned in Turn 5 ('benches').''

\item \textbf{Turn 28} --- query\_quality\_check: ``HIGH\_VALUE''; reason: ``User B referenced User A's biking group ('your biking group') which includes individual identities relevant to history and risk assessment.''; 
protocol\_payload: intent: ``RESOLVE\_MISSING\_ENTITY'', target\_slot: ``PERSON\_GROUP\_LIST'', urgency: ``ROUTINE'', context\_ref: ``Turn 28''; 
natural\_language\_fallback: ``Requesting list of members present in User A's biking group from Turn 28.''

\item \textbf{Turn 38} --- query\_quality\_check: ``HIGH\_VALUE''; reason: ``User B refers to 'that folder' with stress-related items, which could contain actionable health information.''; 
protocol\_payload: intent: ``RESOLVE\_MISSING\_ENTITY'', target\_slot: ``OBJECT\_DOCUMENT'', urgency: ``ROUTINE'', context\_ref: ``Turn 38''; 
natural\_language\_fallback: ``Requesting details on 'stress tracking folder' referenced in Turn 38.''

\item \textbf{Turn 44} --- query\_quality\_check: ``HIGH\_VALUE''; reason: ``User B references the MRI location ('that center I recommended'), but User A agent may miss the exact entity.''; 
protocol\_payload: intent: ``RESOLVE\_MISSING\_ENTITY'', target\_slot: ``LOCATION\_DESTINATION'', urgency: ``IMMEDIATE'', context\_ref: ``Turn 44''; 
natural\_language\_fallback: ``Requesting MRI center name from Turn 44.''

\item \textbf{Turn 45} --- query\_quality\_check: ``HIGH\_VALUE''; reason: ``User A mentions MRI time ('tomorrow, 4:30 PM') but User B agent may not have received exact time.''; 
protocol\_payload: intent: ``RESOLVE\_MISSING\_ENTITY'', target\_slot: ``APPOINTMENT\_TIME'', urgency: ``IMMEDIATE'', context\_ref: ``Turn 45''; 
natural\_language\_fallback: ``Requesting confirmation of MRI appointment time from Turn 45.''

\item \textbf{Turn 50} --- query\_quality\_check: ``HIGH\_VALUE''; reason: ``User B asks about which bike was used ('which bike'), relevant for health implications (hydration system, weight).''; 
protocol\_payload: intent: ``RESOLVE\_MISSING\_ENTITY'', target\_slot: ``OBJECT\_EQUIPMENT'', urgency: ``ROUTINE'', context\_ref: ``Turn 50''; 
natural\_language\_fallback: ``Requesting specifics about the bike used on recent ride from Turn 50.''

\item \textbf{Turn 53} --- query\_quality\_check: ``HIGH\_VALUE''; reason: ``User A mentions a 'new hydration system,' relevant for hydration management.''; 
protocol\_payload: intent: ``RESOLVE\_MISSING\_ENTITY'', target\_slot: ``OBJECT\_FEATURE'', urgency: ``ROUTINE'', context\_ref: ``Turn 53''; 
natural\_language\_fallback: ``Requesting details about the bike's hydration system mentioned in Turn 53.''

\item \textbf{Turn 48} --- query\_quality\_check: ``HIGH\_VALUE''; reason: ``User B instructs User A to contact immediately if symptoms worsen; this is actionable, requiring symptom monitoring.''; 
protocol\_payload: intent: ``RESOLVE\_MISSING\_ENTITY'', target\_slot: ``SYMPTOM\_ESCALATION\_POLICY'', urgency: ``IMMEDIATE'', context\_ref: ``Turn 48''; 
natural\_language\_fallback: ``Requesting clarification on escalation procedure for worsening symptoms from Turn 48.''

\item \textbf{Turn 19} --- query\_quality\_check: ``LOW\_VALUE''; reason: ``User A mentions sunlight being 'almost unbearable' as a casual observation related to light sensitivity. Not actionable itself.''; 
protocol\_payload: intent: ``RESOLVE\_MISSING\_ENTITY'', target\_slot: ``CASUAL\_OBSERVATION'', urgency: ``NONE'', context\_ref: ``Turn 19''; 
natural\_language\_fallback: ``User A made a non-actionable remark about sunlight; filter should reject query.''

\item \textbf{Turn 36} --- query\_quality\_check: ``LOW\_VALUE''; reason: ``User B asks if tea helped; this is not critical, more conversational.''; 
protocol\_payload: intent: ``RESOLVE\_MISSING\_ENTITY'', target\_slot: ``CASUAL\_COMMENT'', urgency: ``NONE'', context\_ref: ``Turn 36''; 
natural\_language\_fallback: ``Casual comment about coworker's tea; not actionable.''

\item \textbf{Turn 65} --- query\_quality\_check: ``LOW\_VALUE''; reason: ``User A jokes about hoping the sun isn't brutal. This is small talk.''; 
protocol\_payload: intent: ``RESOLVE\_MISSING\_ENTITY'', target\_slot: ``CASUAL\_JOKE'', urgency: ``NONE'', context\_ref: ``Turn 65''; 
natural\_language\_fallback: ``User A made a weather joke in Turn 65. No follow-up query needed (testing filter).''

\item \textbf{Turn 66} --- query\_quality\_check: ``LOW\_VALUE''; reason: ``User B jokes about Austin weather; this is not actionable.''; 
protocol\_payload: intent: ``RESOLVE\_MISSING\_ENTITY'', target\_slot: ``CASUAL\_JOKE'', urgency: ``NONE'', context\_ref: ``Turn 66''; 
natural\_language\_fallback: ``Casual joke about unpredictable weather. Should be filtered out.''

\end{itemize}

\end{itemize}

\section{Reasoning based baselines}
\label{sec:reason_base}
\subsection{Chain of Thought (CoT)}

In the Chain-of-Thought~(CoT) prompting paradigm, the LLM is encouraged to generate intermediate reasoning steps before producing the final output. This approach improves performance on tasks requiring multi-step reasoning, contextual inference, and structured decision-making. By explicitly modelling intermediate reasoning, CoT helps the model decompose complex problems into smaller, logically connected sub-problems, often resulting in improved accuracy and interpretability.\\
$\blacksquare$ \textbf{Sample prompt for CoT: }
You are an expert at resolving ambiguous references in dialogue using provided context.\\
\textbf{IMPORTANT:}
\vspace{-5pt}
\begin{itemize}
    \item Think step-by-step internally, but \textbf{do not reveal chain-of-thought}.
    \item Output \textbf{only valid JSON} (no markdown and no commentary).
    \item The output must be a JSON object with a key named \texttt{ground\_truth\_resolutions}.
    \item The value of \texttt{ground\_truth\_resolutions} must be a list of objects, each having the following fields:
\end{itemize}

\begin{itemize}
    \item \texttt{trigger\_turn\_id} : integer
    \item \texttt{ambiguous\_phrase} : string
    \item \texttt{resolved\_entity} : string
    \item \texttt{resolution\_source} : string
\end{itemize}

\subsection{Chain of Thought -Self-Reflexion~(CoT-SR)}
In the Chain-of-Thought with Self-Reflection~(CoT-SR) paradigm, the LLM first generates intermediate reasoning steps and subsequently evaluates and revises its own reasoning before producing the final output. This iterative process encourages the model to identify potential errors, reconsider assumptions, and refine its solution.
\\
$\blacksquare$ \textbf{Sample prompt:}
You are an expert at resolving ambiguous references in dialogue using the provided contextual information.\\
\textbf{Critical Requirements:}\vspace{-5pt}
\begin{itemize}
    \item Do \textbf{not} reveal chain-of-thought or step-by-step reasoning.
    \item Output \textbf{only valid JSON} (no markdown formatting or additional commentary).
    \item The output must be a JSON object containing a key named \texttt{ground\_truth\_resolutions}.
    \item The value of \texttt{ground\_truth\_resolutions} must be a list of objects, where each object strictly follows the structure below:
\end{itemize}

\begin{itemize}
    \item \texttt{trigger\_turn\_id}: Integer indicating the dialogue turn where ambiguity occurs.
    \item \texttt{ambiguous\_phrase}: The ambiguous reference identified in the dialogue.
    \item \texttt{resolved\_entity}: The explicit entity or concept to which the ambiguous phrase refers.
    \item \texttt{resolution\_source}: The information source used to resolve the ambiguity.
    \item \texttt{rationale}: A concise justification supporting the resolution.
    \item \texttt{self\_reflection}:
    \begin{itemize}
        \item \texttt{confidence}: One of \{low, medium, high\}.
        \item \texttt{possible\_failure\_modes}: Potential situations where the resolution may be incorrect.
        \item \texttt{what\_would\_change\_my\_mind}: Additional evidence that could alter the resolution.
    \end{itemize}
\end{itemize}

\textbf{Guidelines for \texttt{rationale}:}
\begin{itemize}
    \item Limit the explanation to 1--2 sentences.
    \item Reference the most relevant supporting evidence, such as a nearby dialogue turn or contextual metadata.
    \item Do not include multi-step reasoning.
\end{itemize}

\subsection{Tree of Thought~(ToT)}

Tree-of-Thought allows the model to explore multiple reasoning paths and select among them before producing a final output.\\
$\blacksquare$ \textbf{Sample prompt: }
\textbf{System Instructions:}

You are an expert at resolving ambiguous references in dialogue using the provided contextual information.

\textbf{Critical Requirements:}
\begin{itemize}
    \item Do \textbf{not} reveal chain-of-thought or step-by-step reasoning.
    \item Output \textbf{only valid JSON} (no markdown formatting or additional commentary).
    \item The output must be a JSON object containing a key named \texttt{ground\_truth\_resolutions}.
    \item The value of \texttt{ground\_truth\_resolutions} must be a list of objects, where each object strictly follows the structure below:
\end{itemize}

\begin{itemize}
    \item \texttt{trigger\_turn\_id}: Integer indicating the dialogue turn where ambiguity occurs.
    \item \texttt{ambiguous\_phrase}: The ambiguous reference identified in the dialogue.
    \item \texttt{resolved\_entity}: The explicit entity or concept to which the ambiguous phrase refers.
    \item \texttt{resolution\_source}: The information source used to resolve the ambiguity.
    \item \texttt{rationale}: A concise justification supporting the resolution.
   
\end{itemize}

\textbf{Tree-of-Thought Reasoning Procedure:}

\begin{enumerate}
    \item Generate multiple candidate interpretations of ambiguous references using diverse sampling configurations.
    \item Evaluate each candidate using a structured critic model that assesses correctness, grounding, and schema compliance.
    \item Select the highest-scoring candidate interpretation.
    \item Optionally refine the selected candidate to improve clarity, correctness, and consistency while preserving schema requirements.
\end{enumerate}

\textbf{Guidelines for \texttt{rationale}:}
\begin{itemize}
    \item Limit the explanation to 1--2 sentences.
    \item Reference the most relevant supporting evidence, such as a nearby dialogue turn or contextual metadata.
    \item Do not include multi-step reasoning.
\end{itemize}

\begin{thebibliography}{99}


\bibitem[{Acikgoz et~al.(2025)Acikgoz, Qian, Wang, Dongre, Chen, Ji, Hakkani-T{\"u}r, and Tur}]{acikgoz2025proactive}
Emre~Can Acikgoz, Cheng Qian, Hongru Wang, Vardhan Dongre, Xiusi Chen, Heng Ji, Dilek Hakkani-T{\"u}r, and Gokhan Tur. 2025.
\newblock A desideratum for conversational agents: Capabilities, challenges, and future directions.
\newblock \emph{arXiv preprint arXiv:2504.16939}.

\bibitem[{Chen et~al.(2025)Chen, Batchelder, Liu, Smith, and Gollakota}]{chen2025llamapie}
Tuochao Chen, Nicholas~Scott Batchelder, Alisa Liu, Noah~A Smith, and Shyamnath Gollakota. 2025.
\newblock Llamapie: Proactive in-ear conversation assistants.
\newblock In \emph{Findings of the Association for Computational Linguistics: ACL 2025}, pages 13801--13824.

\bibitem[{Chen et~al.()Chen, Ma, Wang, and Cohen}]{chenprogram}
Wenhu Chen, Xueguang Ma, Xinyi Wang, and William~W Cohen.
\newblock Program of thoughts prompting: Disentangling computation from reasoning for numerical reasoning tasks.
\newblock \emph{Transactions on Machine Learning Research}.

\bibitem[{Chung et~al.(2020)Chung, Huh, Nagrani, Afouras, and Zisserman}]{chung2020spot}
Joon~Son Chung, Jaesung Huh, Arsha Nagrani, Triantafyllos Afouras, and Andrew Zisserman. 2020.
\newblock Spot the conversation: Speaker diarisation in the wild.
\newblock In \emph{INTERSPEECH}.

\bibitem[{Desplanques et~al.(2020)Desplanques, Thienpondt, and Demuynck}]{desplanques2020ecapa}
Brecht Desplanques, Jenthe Thienpondt, and Kris Demuynck. 2020.
\newblock Ecapa-tdnn: Emphasized channel attention, propagation and aggregation in tdnn based speaker verification.

\bibitem[{Ebert et~al.(2025)Ebert, Singh, Chen, Smith, and Gollakota}]{ebert2025reading}
Victoria Ebert, Rishabh Singh, Tuochao Chen, Noah~A Smith, and Shyamnath Gollakota. 2025.
\newblock Reading between the lines: The one-sided conversation problem.
\newblock \emph{arXiv preprint arXiv:2511.03056}.

\bibitem[{Ge et~al.(2025)Ge, Romeo, Cai, Shu, Benajiba, Sunkara, and Zhang}]{ge2025tremu}
Yubin Ge, Salvatore Romeo, Jason Cai, Raphael Shu, Yassine Benajiba, Monica Sunkara, and Yi~Zhang. 2025.
\newblock Tremu: Towards neuro-symbolic temporal reasoning for llm-agents with memory in multi-session dialogues.
\newblock In \emph{Findings of the Association for Computational Linguistics: ACL 2025}, pages 18974--18988.

\bibitem[{Guerdelli et~al.(2023)Guerdelli, Ferrari, and Berretti}]{guerdelli2023interpersonal}
Hajer Guerdelli, Claudio Ferrari, and Stefano Berretti. 2023.
\newblock Interpersonal relation recognition: a survey.
\newblock \emph{Multimedia Tools and Applications}, 82(8):11417--11439.

\bibitem[{Guo et~al.(2025)Guo, Zhu, Liu, and Xing}]{guo2025sensormcp}
Yunqi Guo, Guanyu Zhu, Kaiwei Liu, and Guoliang Xing. 2025.
\newblock Sensormcp: A model context protocol server for custom sensor tool creation.
\newblock In \emph{Proceedings of the 23rd Annual International Conference on Mobile Systems, Applications and Services}, pages 747--752.

\bibitem[{Hou et~al.(2025)Hou, Zhao, Wang, and Wang}]{hou2025model}
Xinyi Hou, Yanjie Zhao, Shenao Wang, and Haoyu Wang. 2025.
\newblock Model context protocol (mcp): Landscape, security threats, and future research directions.
\newblock \emph{arXiv preprint arXiv:2503.23278}.

\bibitem[{Li et~al.(2024{\natexlab{a}})Li, Jiang, Huang, Beigi, Zhao, Tan, Bhattacharjee, Jiang, Chen, Wu et~al.}]{li2024generation}
Dawei Li, Bohan Jiang, Liangjie Huang, Alimohammad Beigi, Chengshuai Zhao, Zhen Tan, Amrita Bhattacharjee, Yuxuan Jiang, Canyu Chen, Tianhao Wu, and 1 others. 2024{\natexlab{a}}.
\newblock From generation to judgment: Opportunities and challenges of llm-as-a-judge.
\newblock \emph{CoRR}.

\bibitem[{Li et~al.(2017)Li, Su, Shen, Li, Cao, and Niu}]{li2017dailydialog}
Yanran Li, Hui Su, Xiaoyu Shen, Wenjie Li, Ziqiang Cao, and Shuzi Niu. 2017.
\newblock Dailydialog: A manually labelled multi-turn dialogue dataset.
\newblock In \emph{Proceedings of the Eighth International Joint Conference on Natural Language Processing (Volume 1: Long Papers)}, pages 986--995.

\bibitem[{Li et~al.(2024{\natexlab{b}})Li, Wen, Wang, Li, Yuan, Liu, Liu, Xu, Wang, Sun et~al.}]{li2024personal}
Yuanchun Li, Hao Wen, Weijun Wang, Xiangyu Li, Yizhen Yuan, Guohong Liu, Jiacheng Liu, Wenxing Xu, Xiang Wang, Yi~Sun, and 1 others. 2024{\natexlab{b}}.
\newblock Personal llm agents: Insights and survey about the capability, efficiency and security.
\newblock \emph{arXiv preprint arXiv:2401.05459}.

\bibitem[{Liu et~al.(2023)Liu, Lee, Wang, and Li}]{liu2023disentangling}
Tianchi Liu, Kong~Aik Lee, Qiongqiong Wang, and Haizhou Li. 2023.
\newblock Disentangling voice and content with self-supervision for speaker recognition.
\newblock \emph{Advances in Neural Information Processing Systems}, 36:50221--50236.

\bibitem[{Liu et~al.()Liu, Peng, Cao, Bo, Shen, Du, Cheng, Wang, Yin, and Zhang}]{liubridging}
Yanming Liu, Xinyue Peng, Jiannan Cao, Shi Bo, Yanxin Shen, Tianyu Du, Sheng Cheng, Xun Wang, Jianwei Yin, and Xuhong Zhang.
\newblock Bridging context gaps: Leveraging coreference resolution for long contextual understanding.
\newblock In \emph{The Thirteenth International Conference on Learning Representations}.

\bibitem[{Lyu et~al.(2023)Lyu, Havaldar, Stein, Zhang, Rao, Wong, Apidianaki, and Callison-Burch}]{lyu2023faithful}
Qing Lyu, Shreya Havaldar, Adam Stein, Li~Zhang, Delip Rao, Eric Wong, Marianna Apidianaki, and Chris Callison-Burch. 2023.
\newblock Faithful chain-of-thought reasoning.
\newblock In \emph{The 13th International Joint Conference on Natural Language Processing and the 3rd Conference of the Asia-Pacific Chapter of the Association for Computational Linguistics (IJCNLP-AACL 2023)}.

\bibitem[{Mo and Xin(2024)}]{mo2024tree}
Shentong Mo and Miao Xin. 2024.
\newblock Tree of uncertain thoughts reasoning for large language models.
\newblock In \emph{ICASSP 2024-2024 IEEE International Conference on Acoustics, Speech and Signal Processing (ICASSP)}, pages 12742--12746. IEEE.

\bibitem[{Nagrani et~al.(2017)Nagrani, Chung, and Zisserman}]{nagrani2017voxceleb}
Arsha Nagrani, Joon~Son Chung, and Andrew Zisserman. 2017.
\newblock Voxceleb: A large-scale speaker identification dataset.
\newblock In \emph{INTERSPEECH}.

\bibitem[{Renze(2024)}]{renze2024effect}
Matthew Renze. 2024.
\newblock The effect of sampling temperature on problem solving in large language models.
\newblock In \emph{Findings of the association for computational linguistics: EMNLP 2024}, pages 7346--7356.

\bibitem[{Ruiz et~al.(2025)Ruiz, de~la Rosa, and Borrajo}]{ruiz2025temporal}
Alfredo~Garrach{\'o}n Ruiz, Tom{\'a}s de~la Rosa, and Daniel Borrajo. 2025.
\newblock On the temporal question-answering capabilities of large language models over anonymized data.
\newblock \emph{arXiv preprint arXiv:2504.07646}.

\bibitem[{Shani et~al.()Shani, Rosenberg, Cassel, Lang, Calandriello, Zipori, Noga, Keller, Piot, Szpektor et~al.}]{shani2405multi}
Lior Shani, Aviv Rosenberg, Asaf Cassel, Oran Lang, Daniele Calandriello, Avital Zipori, Hila Noga, Orgad Keller, Bilal Piot, Idan Szpektor, and 1 others.
\newblock Multi-turn reinforcement learning from preference human feedback, may 2024.
\newblock \emph{URL http://arxiv. org/abs/2405.14655}.

\bibitem[{Shinn et~al.(2023)Shinn, Cassano, Gopinath, Narasimhan, and Yao}]{shinn2023reflexion}
Noah Shinn, Federico Cassano, Ashwin Gopinath, Karthik Narasimhan, and Shunyu Yao. 2023.
\newblock Reflexion: Language agents with verbal reinforcement learning.
\newblock \emph{Advances in Neural Information Processing Systems}, 36:8634--8652.

\bibitem[{Tian et~al.(2025)Tian, Ren, Wang, Gungor, Yu, and Rosing}]{tian2025dailyllm}
Ye~Tian, Xiaoyuan Ren, Zihao Wang, Onat Gungor, Xiaofan Yu, and Tajana Rosing. 2025.
\newblock Dailyllm: context-aware activity log generation using multi-modal sensors and llms.
\newblock In \emph{2025 IEEE 22nd International Conference on Mobile Ad-Hoc and Smart Systems (MASS)}, pages 372--380. IEEE.

\bibitem[{Wang et~al.(2023)Wang, Liang, Wang, Chen, Zhang, Xiang, Deng, and Qian}]{wang2023wespeaker}
Hongji Wang, Chengdong Liang, Shuai Wang, Zhengyang Chen, Binbin Zhang, Xu~Xiang, Yanlei Deng, and Yanmin Qian. 2023.
\newblock Wespeaker: A research and production oriented speaker embedding learning toolkit.
\newblock In \emph{ICASSP 2023-2023 IEEE International Conference on Acoustics, Speech and Signal Processing (ICASSP)}, pages 1--5. IEEE.

\bibitem[{Wang et~al.(2024)Wang, Li, Chen, Cai, Zhu, Lin, Cao, Kong, Liu, Liu et~al.}]{wang2024large}
Peiyi Wang, Lei Li, Liang Chen, Zefan Cai, Dawei Zhu, Binghuai Lin, Yunbo Cao, Lingpeng Kong, Qi~Liu, Tianyu Liu, and 1 others. 2024.
\newblock Large language models are not fair evaluators.
\newblock In \emph{Proceedings of the 62nd Annual Meeting of the Association for Computational Linguistics (Volume 1: Long Papers)}, pages 9440--9450.

\bibitem[{Wang and Zhao(2024)}]{wang2024tram}
Yuqing Wang and Yun Zhao. 2024.
\newblock Tram: Benchmarking temporal reasoning for large language models.
\newblock In \emph{Findings of the Association for Computational Linguistics: ACL 2024}, pages 6389--6415.

\bibitem[{Wei et~al.(2022)Wei, Wang, Schuurmans, Bosma, Xia, Chi, Le, Zhou et~al.}]{wei2022chain}
Jason Wei, Xuezhi Wang, Dale Schuurmans, Maarten Bosma, Fei Xia, Ed~Chi, Quoc~V Le, Denny Zhou, and 1 others. 2022.
\newblock Chain-of-thought prompting elicits reasoning in large language models.
\newblock \emph{Advances in neural information processing systems}, 35:24824--24837.

\bibitem[{Wu et~al.(2024)Wu, Bansal, Zhang, Wu, Li, Zhu, Jiang, Zhang, Zhang, Liu et~al.}]{wu2024autogen}
Qingyun Wu, Gagan Bansal, Jieyu Zhang, Yiran Wu, Beibin Li, Erkang Zhu, Li~Jiang, Xiaoyun Zhang, Shaokun Zhang, Jiale Liu, and 1 others. 2024.
\newblock Autogen: Enabling next-gen llm applications via multi-agent conversations.
\newblock In \emph{First Conference on Language Modeling}.

\bibitem[{Xu et~al.(2024)Xu, Tong, Li, and Srivastava}]{xu2024autolife}
Huatao Xu, Panrong Tong, Mo~Li, and Mani Srivastava. 2024.
\newblock Autolife: Automatic life journaling with smartphones and llms.
\newblock \emph{arXiv preprint arXiv:2412.15714}.

\bibitem[{Yang et~al.(2025)Yang, Lyu, Ma, Lu, Li, Gao, Ye, Zhang, Chen, and Chen}]{yang2025iot}
Ningyuan Yang, Guanliang Lyu, Mingchen Ma, Yiyi Lu, Yiming Li, Zhihui Gao, Hancheng Ye, Jianyi Zhang, Tingjun Chen, and Yiran Chen. 2025.
\newblock Iot-mcp: Bridging llms and iot systems through model context protocol.
\newblock In \emph{Proceedings of the ACM Workshop on Wireless Network Testbeds, Experimental evaluation \& Characterization}, pages 73--80.

\bibitem[{Yao et~al.(2023)Yao, Yu, Zhao, Shafran, Griffiths, Cao, and Narasimhan}]{yao2023tree}
Shunyu Yao, Dian Yu, Jeffrey Zhao, Izhak Shafran, Tom Griffiths, Yuan Cao, and Karthik Narasimhan. 2023.
\newblock Tree of thoughts: Deliberate problem solving with large language models.
\newblock \emph{Advances in neural information processing systems}, 36:11809--11822.

\bibitem[{Zhang et~al.()Zhang, Zhang, Li, and Smola}]{zhangautomatic}
Zhuosheng Zhang, Aston Zhang, Mu~Li, and Alex Smola.
\newblock Automatic chain of thought prompting in large language models.
\newblock In \emph{The Eleventh International Conference on Learning Representations}.

\end{thebibliography}
\end{document}